\documentclass[11pt]{article}

\usepackage[]{acl}

\usepackage{times}
\usepackage{latexsym}
\usepackage{multirow}
\usepackage{graphicx}
\usepackage{enumitem}
\usepackage[linesnumbered,ruled,vlined]{algorithm2e}
\usepackage{amsmath}
\usepackage{amssymb}
\usepackage{latexsym}
\usepackage[inkscapelatex=false]{svg}
\usepackage{url}

\SetKwInput{KwInput}{Input}                
\SetKwInput{KwOutput}{Output}              

\usepackage{makecell}
\usepackage{color}
\usepackage{booktabs}
\usepackage{hyperref}
\usepackage{float}
\usepackage[all]{nowidow}
\usepackage{CJKutf8}    
\usepackage[T1]{fontenc}
\usepackage{microtype}
 
\title{Improving Chinese Story Generation via Awareness of Syntactic Dependencies and Semantics}

\author{Henglin Huang\textsuperscript{1}\footnotemark[1], Chen Tang\textsuperscript{1}\footnotemark[1], Tyler Loakman\textsuperscript{2}, Frank Guerin\textsuperscript{1} and Chenghua Lin\textsuperscript{2}\footnotemark[2]\\
  \textsuperscript{1}Department of Computer Science, The University of Surrey, UK \\
  \textsuperscript{2}Department of Computer Science, The University of Sheffield, UK \\
  \texttt{\{hh01034,chen.tang,f.guerin\}@surrey.ac.uk} \\
  \texttt{\{tcloakman1,c.lin\}@sheffield.ac.uk}}

\begin{document}
\begin{CJK}{UTF8}{gbsn}

\maketitle

\renewcommand{\thefootnote}{\fnsymbol{footnote}} 
\footnotetext[1]{Equal contribution.} 
\footnotetext[2]{Corresponding author.}

\renewcommand*{\thefootnote}{\arabic{footnote}}

\begin{abstract}
Story generation aims to generate a long narrative conditioned on a given input. In spite of the success of prior works with the application of pre-trained models, current neural models for Chinese stories still struggle to generate high-quality long text narratives. We hypothesise that this stems from ambiguity in syntactically parsing the Chinese language, which does not have explicit delimiters for word segmentation. Consequently, neural models suffer from the inefficient capturing of features in Chinese narratives. 
In this paper, we present a new generation framework that enhances the feature capturing mechanism by informing the generation model of dependencies between words and additionally augmenting the semantic representation learning through synonym denoising training. We conduct a range of experiments, and the results demonstrate that our framework outperforms the state-of-the-art Chinese generation models on all evaluation metrics, demonstrating the benefits of enhanced dependency and semantic representation learning.

\end{abstract}

\section{Introduction}
Story generation presents a challenging task, primarily due to the difficulty that end-to-end neural models experience in maintaining logical coherence during long text generation \cite{tang2022recent}. These challenges are even more prominent for the task of story generation in Chinese, due to increased levels of ambiguity stemming from the absence of explicit delimiters for character separation \cite{tian-etal-2020-improving-chinese}. Recent works, on the other hand, have aimed to improve long text generation through the proposal of more efficient frameworks \cite{rashkin-etal-2020-plotmachines, goldfarb-tarrant-etal-2020-content,brahman-chaturvedi-2020-modeling}, or through augmenting existing frameworks with pre-training and the injection of extra knowledge \cite{xu-etal-2020-megatron, guan-etal-2020-knowledge,guan-etal-2022-lot}.

\begin{figure}[ht]
\centering
\includegraphics[width=0.9\columnwidth]{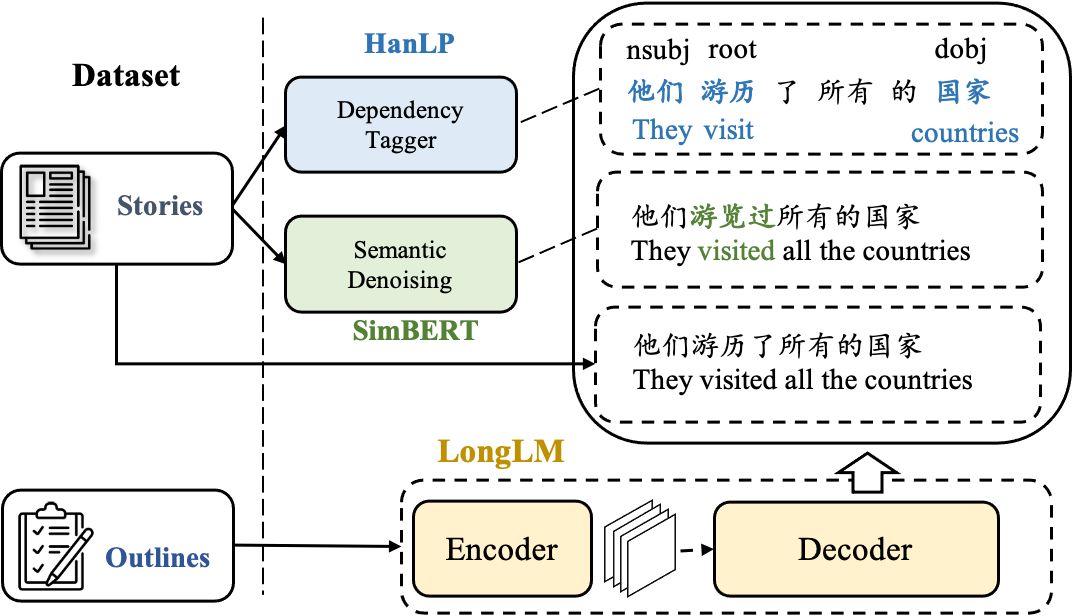}
\caption{The overview of our framework. For the stories, the words in blue denote the semantic roles in a sentence (e.g., the subject (nsubj)), whilst the words in green denote the expressions that are replaced with synonyms.} 
\label{fig:overview}
\end{figure}

However, we observe that current generation models still struggle to generate fluent and coherent Chinese stories, which may be the result of the inefficient capturing of features in written Chinese. For example, Chinese characters have a range of morphological parsing strategies, e.g., “ 小心地滑” can be understood as “小心 地滑” (caution wet floor) or “小心地 滑” (carefully slide), whose meaning is highly dependent on context~\cite{chen-etal-2018-simplenlg,Li-etal-2018-metaGen}. This may cause important sentential roles such as subjects, predicates, and objects to be difficult to identify and process by neural models. Additionally, when neural networks learn the semantics of an utterance, synonymous expressions may lead to confusion, damaging the robustness of the generation model, e.g., “游历”, “周游”, and “游览” are different Chinese words but all express "travelling" in Chinese. We therefore propose to train neural networks to learn the semantic-level features contained in context, rather than the low-level features of characters.

To this end, we propose a novel data augmented story generation framework illustrated in \autoref{fig:overview}, including a LongLM \cite{guan-etal-2022-lot} based conditional generator, a dependency tagger, and a semantic denoising module. 
The generator, LongLM \cite{guan-etal-2022-lot}, is a SOTA pre-trained model that has been demonstrated to be effective at multiple Chinese NLG tasks. The dependency tagger, powered by HanLP\footnote{\url{https://github.com/hankcs/HanLP}} \cite{he-choi-2021-stem}, recognises the root of a sentence, usually the verb, as well as related subjects and objects via dependency parsing, all of which are essential in expressing the event represented within a sentence. The semantic denoising module, based on SimBert\footnote{\url{https://github.com/ZhuiyiTechnology/simbert}} \cite{su2020simbert}, generates a range of different, yet essentially synonymous sentences, to force the neural network learn the semantic representations of key entities and different expressions. Overall, our proposed framework enhances the ability for language understanding in written Chinese via training to capture the dependencies and semantics contained within sentences, in order to then generate stories.

We conduct a range of experiments on the latest public benchmark for Chinese story generation \cite{guan-etal-2022-lot}, and the results indicate that the model trained with our framework substantially outperforms the state-of-the-art (SOTA) baselines on all metrics.\footnote{Our code for reproduction is available at \url{https://github.com/hehedaozuiteng/Chinese-Story-Generation}.} This indicates that our framework improves the generated stories via enhanced capturing of syntactic dependencies and semantic features.

\section{Methodology}

We formulate our story generation task based on the OutGen task from LOT \cite{guan-etal-2022-lot}, a Chinese story generation benchmark. The definition of the task is: An outline $X$, which contains an unordered list of an arbitrary number of Chinese phrases concerning characters and events, is given as the input. The model is required to generate a coherent story $Y = \{y_1, y_2, ..., y_n\} $ where  $y_i$ denotes the $ i $-th token (Chinese character) in the story.

\subsection{Dependency Tagging}
We employ HanLP \cite{he-choi-2021-stem} to parse dependencies within Chinese stories. Unlike in English, the basic unit of Chinese dependency parsing is the word segment denoted as $\mathit{Seg} = \{\mathrm{token}_1, ..., \mathrm{token}_m\} $, which contains $m$ tokens. Therefore, a story can be represented as $Y = \{\mathit{Seg}_1, ...\}$. For each story, we firstly identify the set of dependencies $\mathit{T} = \{\mathit{Seg}_{h}, \mathit{D}_{\mathit{tag}}, \mathit{Seg}_{t}\}$, and then select target labels $\mathit{T}_{\mathit{target}}$ to insert into the original stories. These target labels are $\mathit{nsubj}$ (representing subjects), $\mathit{root}$ (usually representing verbs), $\mathit{dobj}$ (representing direct objects), and $\mathit{pobj}$ (representing indirect objects following prepositions) \cite{de2008stanford}. The process is depicted as below:
\begin{align}
    & \mathit{T}_{\mathit{target}} = \mathit{D}_{\mathit{\mathit{tag} \in \{\mathit{nsubj}, \mathit{root}, \mathit{dobj}, \mathit{pobj}\}}} \\
    & \mathit{Tagger}(\mathit{Seg}_i) = \left\{
                    \begin{array}{ll}
                    \mathit{Seg}_i,\mathit{D}_{\mathit{tag}} &  {  \mathit{D}_{\mathit{tag}} \in \mathit{T}_{\mathit{target}}}\\
                    \mathit{Seg}_i       &   \mathit{otherwise}
                    \end{array} \right. \\
    & Y^{\mathit{D}} = \mathit{Tagger}(Y, \mathit{T}_{\mathit{target}}) 
\end{align}
where $Y^{\mathit{D}}$ is a story with target dependency labels. For instance, the input “他们 游历 了 所有 的 国家” ("They visited all the countries") will be tagged, and the output would be “他们<nsubj> 游历<root> 了 所有 的 国家<dobj>” (They<nsubj> visited<root> all the countries<dobj>).

\begin{table*}[!t]
\centering
\resizebox{\linewidth}{!}{
\begin{tabular}{l|ccccccc|ccccccc}
\toprule[1pt]
\multirow{2}{*}{\textbf{Methods}} & \multicolumn{7}{c|}{\textbf{Validation Set}} & \multicolumn{7}{c}{\textbf{Test Set}}\\
    & \textbf{B-1} & \textbf{B-2} & \textbf{D-1} & \textbf{D-2} & \textbf{cover}          
  & \textbf{order} &  \textbf{Overall}
  & \textbf{B-1} & \textbf{B-2} & \textbf{D-1} & \textbf{D-2} & \textbf{cover} 
  & \textbf{order} &\textbf{Overall} \\
\hline
\textbf{ConvS2S} &29.23&10.38&3.45&21.79&14.81&25.34&16.08 &29.00& 10.14& 1.60& 13.95& 15.45& 25.77&15.19 \\
\textbf{Fusion} &29.22 &10.34& 3.39& 22.67& 17.41& 26.55&16.5 &28.77 &10.22& 1.47& 14.12& 17.10& 26.36&15.40\\
\midrule
\textbf{GPT2}$_\mathit{base}$ &30.43 &14.87 &10.95 &44.38 &60.90 &55.52 &28.43 &30.17 &14.91 &7.62 &36.87 &60.87 &55.90 &27.62 \\
\textbf{GPT2}$_\mathit{base}^{\dagger}$ & 35.29 &18.31& 13.89& 51.36& 64.01& 57.64&32.26&35.79&18.68&9.89&43.52&64.43&56.96&31.57\\
\textbf{PM} &31.81& 14.94& 12.99& 50.56& 62.98& 56.75&29.87&31.85&15.24&8.62&41.32&63.15&57.21&28.99\\
\textbf{PW} &35.84& 18.47& 11.86& 47.62& 64.93& 57.30 &31.89 & 35.12 & 17.96 & 8.68 &40.17 &63.70& 55.17& 30.44\\
\textbf{mT5}$_\mathit{base}$ &36.71 &22.25 &14.52 &50.01 &77.98 &63.15& 35.93 &36.33 &22.07 &10.90 &43.65& 78.66& 63.79&35.19\\
\textbf{LongLM}$_\mathit{base}$  &40.33 &24.29 &14.66 &51.82 &79.60 &62.78 &37.75      &40.25 &24.15 &10.75 &44.40 &79.88 &63.67 &36.92 \\
\textbf{LongLM}$_\mathit{large}$ &42.79 &24.91 &16.13 &57.71 &80.46 &64.36 &39.44 &42.10 &24.77 &12.04& 50.29& 81.48& 64.82 &38.53\\
\midrule
\textbf{Ours} & \textbf{44.40} & \textbf{25.49} &\textbf{17.35} &\textbf{62.47} &\textbf{88.93} &\textbf{64.72} &\textbf{41.41} & \textbf{44.82} &\textbf{25.88} &\textbf{12.31} &\textbf{53.21} &\textbf{89.15} &\textbf{67.05} & \textbf{40.78} \\
\midrule
metric weight \textit{w}$_\mathit{i}$ & \textit{0.190} & \textit{0.405} & \textit{0.119} & \textit{0.095} & \textit{0.095} & \textit{0.095} & \textit{0.999}
& \textit{0.195} & \textit{0.390} & \textit{0.122} & \textit{0.098} & \textit{0.098} & \textit{0.098} & \textit{1.00} \\
Reference & \textit{100.00} & \textit{100.00} & \textit{21.66} & \textit{71.43} & \textit{100.00} & \textit{100.00} & \textit{92.23}
& \textit{100.00} & \textit{100.00} & \textit{15.71} & \textit{63.46} & \textit{100.00} & \textit{100.00} & \textit{91.64}
\\
\bottomrule[1pt]
\end{tabular}
}
\caption{Automatic evaluation of generated stories. The best score on each metric is highlighted in bold. \textit{w}$_\mathit{i}$ is the weight of each metric computed for the overall aggregate score. For all metrics, the higher the score, the better.}
\label{tab:baselines}
\end{table*}

\subsection{Semantic Denoising}
To help neural networks understand the semantics of Chinese segments implicitly contained in sentences, we employ $\mathit{SimBERT}$ \cite{su2020simbert}, which inputs a sentence, and outputs a similar sentence with the same meaning in order to generate a training corpus with large number of synonymous sentences. We therefore aim to train neural networks to resist the semantic noise introduced by different Chinese expressions. For instance, the compound words "去过" and "去了" both represent the meaning "went" in Chinese, in which “去” (go), with different auxiliary characters, may have the same meaning. As this phenomenon is ubiquitous in Chinese, we force the neural networks to denoise the changes in surface forms in order to better understand the semantics of these segments. Consequently, we obtain an augmented data corpus for semantic denoising:
\begin{align}
    & \{..., \mathit{Seg}_i^\prime, ...\} = \mathit{SimBERT}(\{..., \mathit{Seg}_i, ...\}) \\
    & \underbrace{\{Y^{\mathit{S}}_1, ..., Y^{\mathit{S}}_6\}}_{6} = \mathit{SimBERT}(Y) 
\end{align}
where $\mathit{Seg}_i^\prime$ is a synonym of $\mathit{Seg}_i$.  $Y^S$ is a story that is different from $Y$ but has the same input $X$. We generate 6 similar stories for each $X$, and train our neural generator on the enlarged corpus. 

\subsection{Neural Generator}
We employ LongLM \cite{guan-etal-2022-lot}, a Chinese long text pre-trained language model, as the base generator of our framework. It consists of Transformer-based neural blocks \cite{vaswani2017attention,zeng-etal-2021-affective} with an encoder-decoder architecture to generate narratives. The training process is as follows:
\begin{align}
    F &= \mathit{Encoder}(X)\\
    Tagger(\{Y, Y_1^S, ...\}) & \stackrel{predict}{\Longleftarrow} \mathit{Decoder}(F)
\end{align}
where the maximum sequence length is set to 512 for both the $\mathit{Encoder}$ and $\mathit{Decoder}$. LongLM is then fine-tuned with standard cross-entropy loss.

\section{Experiment}

\subsection{Experiment Setup}
\textbf{Dataset} We conduct our experiments on the OutGen task of LOT \cite{guan-etal-2022-lot}, a Chinese story benchmark which consists of 2427 high-quality filtered Chinese stories. Each input outline contains a sequence of $8$ unordered phrases (i.e., their order does not necessarily reflect the order in which they would be present within a narrative). We follow the data split from the benchmark of 60/10/30 for training/validation/testing, respectively. The statistics are shown in \autoref{tab:data_stat}.


\begin{table}[htbp]
\scriptsize
    \centering
    \begin{tabular}{l|ccc}
    \toprule
\textbf{Datasets}&\textbf{Train}&\textbf{Val}&\textbf{Test}\\
\midrule
\textbf{\# Examples}& 1,456 & 242 & 729\\
\textbf{Vocabulary Size} & 19k  & 6k & 12k\\
\midrule
\textbf{Avg. \# Word in Input Title} & 4.64&4.89&4.64\\
\textbf{Avg. \# Word in Input Outline} & 19.20  & 19.05 & 19.47\\
\textbf{Avg. \# Phrase in Input Outline} & 8.00  & 8.00 & 8.00\\
\midrule
\textbf{Avg. \# Char in Output Text}&169.94&169.80&170.49\\
\textbf{Avg. \# Word in Output Text} & 108.91  & 108.68 & 109.04\\
\textbf{Avg. \# Sent in Output Text} & 7.20  & 7.11 & 7.15\\
\bottomrule
    \end{tabular}
    \caption{Data statistics of the OutGen task in LOT. The abbreviations \textbf{char}/\textbf{sent}/\textbf{len} stand for \textbf{character}/\textbf{sentence}/\textbf{length}, respectively.}
    \label{tab:data_stat}
\end{table}


\subsection{Baselines}
We compare our generation framework with a selection of competitive baselines, including the non-pretrained models \textbf{ConvS2S} \cite{gehring2017convolutional} and \textbf{Fusion} \cite{fan2018hierarchical}; pre-trained GPT2 models including \textbf{GPT2$_{base}$} \cite{zhao2019uer} and \textbf{GPT2$^\dagger_{base}$} (the latter of which is pretrained on the benchmark corpus) \cite{guan-etal-2022-lot}); PlotMachines~(\textbf{PM})~\cite{rashkin2020plotmachines}; {Plan\&Write}~(\textbf{PW})~\cite{yao2019plan}; and \textbf{mT5} (based on google/mt5-base) \cite{xue2021mt5}. Specifically, the pre-trained models of baselines are implemented and restored from the prior works on the Chinese language. GPT2 based models are based on uer/gpt2-
chinese-cluecorpussmall \cite{zhao-etal-2019-uer}.

\subsection{Implementation Details}
We restore the publicly available checkpoint\footnote{\url{https://huggingface.co/thu-coai/LongLM-base}} from Huggingface,  and fine-tune LongLM$_\mathit{base}$ within our framework. LongLM has 12 attention heads and 12 hidden layers in each encoder and decoder, leading to a total of 223M parameters. We set the maximum sequence length to 512, the batch size to 3, and use a linear schedule to set the warm up step to 100 and the learning rate to 0.0001 for the Adam optimiser. All models are fine-tuned on 2 Nvidia RTX A5000 GPUs. 

\begin{table*}[htb]
\centering
\resizebox{\linewidth}{!}{
\begin{tabular}{l|ccccccc|ccccccc}
\toprule[1pt]
\multirow{2}{*}{\textbf{Methods}} & \multicolumn{7}{c|}{\textbf{Validation Set}} & \multicolumn{7}{c}{\textbf{Test Set}}\\
    & \textbf{B-1} & \textbf{B-2} & \textbf{D-1} & \textbf{D-2} & \textbf{cover}          
  & \textbf{order} &  \textbf{Overall}
  & \textbf{B-1} & \textbf{B-2} & \textbf{D-1} & \textbf{D-2} & \textbf{cover} 
  & \textbf{order} &\textbf{Overall} \\
\hline
\textbf{LongLM}$_\mathit{base}$  &40.33 &24.29 &14.66 &51.82 &79.60 &62.78 &37.75      &40.25 &24.15 &10.75 &44.40 &79.88 &63.67 &36.92 \\
\midrule
\textbf{\textbf{- w/ Dependencies (D)}} &42.33 &25.08 &15.21 &58.23 & 88.48 & 65.24 &40.21    &42.41 & 25.08 & 11.11&49.69 &89.24&65.21 & 39.33\\
\textbf{\textbf{- w/ Semantics (S)}}  & 41.77&25.78&14.24&57.55&\textbf{89.80}&65.13&40.32& 41.16&25.33&10.25&48.88&\textbf{90.27}&66.25&39.20 \\ 
\textbf{\textbf{- w/ D + S (ours)}}  & \textbf{44.89}&\textbf{25.80}&\textbf{17.13}&\textbf{63.02}&89.06&\textbf{65.55}&\textbf{41.76}&\textbf{44.55}&\textbf{25.70}&\textbf{12.46}&\textbf{53.71}&89.18&\textbf{66.84}&\textbf{40.70} \\ 
\bottomrule[1pt]
\end{tabular}
}
\caption{Automatic evaluation for the ablation study. \textbf{Dependencies} denotes the Dependency Tagging module, and \textbf{Semantics} denotes the Semantic Denoising module.}
\label{tab:ablation}
\end{table*}

\subsection{Evaluation Metrics}
Following the LOT benchmark \cite{guan-etal-2022-lot}, we perform automatic evaluation on the metrics of BLEU-n (\textbf{B-n}) \cite{papineni2002bleu}, Distinct-n (\textbf{D-n}) \cite{DBLP:journals/corr/LiGBGD15}, Coverage (\textbf{cover}), and Order (\textbf{order}). The BLEU-n score measures the quality of generated text by comparing the degree of n-gram overlap with the ground-truth texts; the Distinct score measures the n-gram diversity of the generated text; Coverage is the same as ROUGE-L \cite{lin-2004-rouge}, which measures the recall rate between generated text and input outline phrases; and Order measures the difference between the positional orders of the input phrases in the generated text and the ground-truth text (which is calculated by dividing the number of positional order inversions in the generated story by the number of position pairs between any two phrases) \cite{guan-etal-2022-lot}. We compute the overall aggregate score with the metric weighting scheme presented in LOT.

\subsection{Experimental Result}
\paragraph{Comparison with Baselines}
As shown in \autoref{tab:baselines}, our proposed model substantially outperforms all competitive baselines by a considerable margin. We implement LongLM$_{\mathit{base}}$ (223M hyper-parameters) as our conditional generator. However, the results indicate our model can also significantly outperform LongLM$_{\mathit{large}}$ (1B hyper-parameters), on all metrics. Compared to the SOTA model (LongLM$_{\mathit{large}}$), our proposed model achieves up to a 10\% improvement on several metrics for both the validation and test sets, and around 5\% for the overall score. Additionally, when compared to LongLM$_{\mathit{base}}$, our model demonstrates a performance uplift of around 10\% on the overall score.

The performance improvements seen on BLEU-n and Coverage indicate that our generated stories have a higher degree of overlap with the reference stories. Considering the input outline is unordered, this indicates that via the awareness of dependencies and semantics, our proposed model can better leverage syntactic features, and generate more fluent narratives as a result. The scores on Order (computed by the order of outlines in the generated stories compared to the reference), further demonstrate the improvement on language discourse. Meanwhile, the diversity of stories is also substantially raised, for which we argue that semantic denoising contributes significantly.

Considering the results as a whole, the significant improvements of our model over existing baslines demonstrates that the enhanced capturing of dependencies and semantics contribute to the language understanding task. This is particularly apparent for Chinese, where expressions are more ambiguous due to the lack of explicit delimiters. Using this increased level of understanding, conditional generators can therefore generate more fluent and diverse stories.

\paragraph{Ablation Study}
We conduct ablation experiments presented in \autoref{tab:ablation} to analyse the individual contributions of each module. We observe that the enhanced feature capturing of both the dependencies and semantics substantially improves on the original neural generator, and combining both approaches further improves performance. This indicates that these two features largely perform different functions that contribute to language generation. Whilst our proposed model outperforms all ablated models when considering most  metrics, performance of a single module on some metrics is still close to or even slightly better than the combined model (e.g., on coverage). This phenomenon implies that the two proposed modules may have a small degree of shared function when exploiting features from text. In addition, insufficient training may also lead to the inadequacy of incorporating both features for decoding. We leave further study of incorporating both features to future work.

\paragraph{Case Study}
Several generated Chinese stories are presented in \autoref{apx:case} to further demonstrate the effectiveness of our framework.
\section{Conclusion}
We propose a novel story generation framework for Chinese, which includes a dependency tagging module, a semantic denoising module, and a neural conditional generator. We aim to improve the generation of Chinese through more effectively incorporating the features of dependencies and semantics. The performance improvements shown in our experiments and  ablation study demonstrate that these features significantly benefit the task of Chinese story generation.

\section*{Acknowledgements}
Chen Tang is supported by the China Scholarship Council (CSC) for his doctoral study (File No.202006120039). Tyler Loakman is supported by the Centre for Doctoral Training in Speech and Language Technologies (SLT) and their Applications funded by UK Research and Innovation [grant number EP/S023062/1]. We also gratefully acknowledge the anonymous reviewers for their insightful comments.

\bibliography{bibs/sec1-introduction,
              bibs/sec2-methodology,
              bibs/sec3-experiment}

\appendix
\section{Appendix}
\label{sec:appendix}

\subsection{Case Study} \label{apx:case}
In \autoref{tab:example} we present an example for the basis of a case study. \autoref{tab:case_study} presents the generated stories from the neural generation models, including the SOTA baseline \textbf{LongLM}$_\mathit{base}$, our proposed framework, and its ablated models.

Firstly, with large-scale pre-training on narrative corpora, the generated stories have relatively less repetition and diversity problems than traditional text generation methods. The main issues are now located in linguistic aspects such as fluency, coherence, and relevance. It can be observed that the generated story from the SOTA baseline model suffers from the ambiguity of the Chinese language, which leads to grammatical and semantic errors. For instance, the sentence ``从前，有个挑水夫，他把路旁撒的半桶水送到主人家的破桶留意路旁'' (Once upon a time, there was a water-carrier who sent half a bucket of water sprinkled by the roadside to
the broken bucket at the master’s house keep an eye on the roadside) has grammatical errors. This may result from inadequate  understanding of the dependency roles of each part of the sentences, which leads to misusing two verb phrases ("sent", "keep and eye on"). For the same reason, the linguistic ambiguity makes the model struggle to capture the semantic meaning of each sentence constituent. For example, the sentence ``结果，路旁就完好无损了'' (As a result, the roadside was intact) contains no grammatical errors, but also makes no sense to the story. It can be intuitively supposed that the key words "the roadside" and "intact" in the given outline are directly composed here by the neural model without understanding their semantics.

\begin{table*}[ht]
\begin{center}
\begin{tabular}{p{0.92\linewidth} } 
\toprule[1pt]
\textbf{Outline:} \quad \small "破桶留意路旁", "只能剩下半桶水", "水送到主人家", "挑水夫道歉", "路旁撒", "挑水夫说", "趟挑运", "完好无损" \\
"the broken bucket keeps an eye on the roadside", "only half a bucket of water is left", "deliver water to the master's house", "the water-bearer apologises", "sprinkled by the roadside", "the water-bearer said", "travel to pick up", "intact"\\
\midrule[1pt]
\textbf{Reference Story:}  \quad  \small 挑水夫有两个水桶，一个桶有裂缝，另一个完好无损。每趟挑运之后，好桶总是能将满满一桶水送到主人家中，但是破桶却只能剩下半桶水。破桶非常羞愧。一天，它对挑水夫道歉。挑水夫并没有生气，他让破桶留意路旁的花朵。他们走在山坡上，破桶看到缤纷的花朵，开满在路的一旁。挑水夫说，只有破桶的那一边有花，好桶的那一边却没有。原来挑水夫知道破桶的缺陷，因此善加利用，在破桶那边的路旁撒了花种，每回从溪边过来，破桶就替它一路浇了花。如果不是因为破桶，主人的桌上也没有那么好看的花朵了。 \\ 
The water-bearer had two buckets. One bucket is broken and the another is intact. After each pick-up, the good bucket can always deliver a full bucket of water to the master's house, but the broken bucket only has half a bucket of water left. The broken bucket feels very ashamed. One day, it apologised to the water bearer. The water-bearer was not angry, he told the broken bucket to keep an eye on the flowers by the roadside. As they walked down the hillside, Broken bucket saw colorful flowers that filled the side of the road. The water-bearer said that there were flowers only on the side of the broken bucket, but not on the side of the good bucket. It turned out that the water-bearer knew about the defects of the broken bucket, so he made good use of it.  Water-bearer sowed flower seeds on the roadside over the broken bucket. Every time he came from the stream, the broken bucket would water the flowers along the way. If it weren't for the broken bucket, there would not be such beautiful flowers on the master's table. \\
\bottomrule[1pt]
\end{tabular}
\caption{An example of an outline and its reference story.}
\label{tab:example}
\end{center}
\end{table*}

Regarding the ablation study, similar errors can also be observed in each ablated model: (1) ``主人让他把破桶留意路旁撒的水
送去主人家。'' (The master asked him to send the water the broken bucket noticed sprinkled by the roadside to the master’s house.), in the story generated by \textbf{\textbf{- w/ Dependencies}}, also has obvious grammatical errors; (2) \textbf{\textbf{- w/ Semantics}} generates the sentence ``主人对他道歉，并把那桶水送
到了路旁撒了一次又一次网。'' (The master apologised
to him and sent the bucket of water to the roadside for casting the net again and again.), which is free of grammatical errors, but does not make sense semantically. 

Furthermore, when comparing sentences containing the same key words from outlines in different generated stories, it can be observed that our proposed techniques for dependency parsing and semantic denoising indeed improve the quality of generated sentences. For 
example, for the key word "水送到主人家" (deliver water to
the master’s house), the aforementioned \textbf{LongLM}$_\mathit{base}$ and \textbf{\textbf{- w/ Dependencies}} all generate a sentence with grammatical errors. However, the sentences ``挑水人问他为什么不把水送到主人家'' (The water-bearer
asked him why he didn’t deliver the water to the master’s house.) by \textbf{\textbf{- w/ Semantics}} and ``挑水的人觉得，这趟挑运真是完好无损。'' (Water-
bearer felt that the trip was really intact.) by \textbf{Ours}, are all free of grammatical errors. 

Compared to other generated stories, the story from our proposed model contains less errors, and reads more fluently and coherently. Therefore, with the analysis of these errors, we can confirm that current neural models are likely to suffer from the challenge of parsing dependencies and semantics in Chinese narratives. However, our proposed generative framework can improve this uncertainty brought by language ambiguity, and consequently generates a better story.

\begin{table*}[htbp]
\begin{center}
\begin{tabular}{p{0.92\linewidth} } 
\toprule[1pt]
\hline
\textbf{Outline:} \quad \small "破桶留意路旁", "只能剩下半桶水", "水送到主人家", "挑水夫道歉", "路旁撒", "挑水夫说", "趟挑运", "完好无损" \\
"the broken bucket keeps an eye on the roadside", "Only half a bucket of water is left", "deliver water to the master's house", "The water-bearer apologises", "Sprinkled by the roadside", "the water-bearer said", "travel to pick up", "intact"\\
\midrule[1pt]
\textbf{LongLM}$_\mathit{base}$ \quad \small 从前，有个挑水夫，他把路旁撒的半桶水送到主人家的破桶留意路旁。主人很生气，就把水送到路旁，还把水送了上去。结果，路旁就完好无损了。这下，挑水老大不高兴了。他向主人道歉，说自己没有做好事。主人听后，觉得这件事是个错误，只能剩下半桶的水了。于是，他决定把这趟挑运的收获全部归到主人的头上。从此，主人再也不敢欺负他了，再也没有欺负过他了。
\\
Once upon a time, there was a water-bearer who sent half a bucket of water sprinkled by the roadside to the master's house and the broken bucket kept an eye on the roadside. The master was very angry, so he sent the water to the side of the road, and also sent the water up. As a result, the roadside was intact. This time, the water-bearer boss was not happy. He apologised to his master and said that he had done nothing good. When the master heard this, the master felt that the matter was a mistake and only half a bucket of water could be left. Therefore, he decided to attribute all the harvest of this fortune to the master. From then on, the master never dared to bully him again, never bullied him again. \\
\midrule
\textbf{\textbf{- w/ Dependencies}} \quad \small 从前，有个挑水夫，他想把路旁撒的半桶水送到主人家。主人让他把破桶留意路旁撒的水送去主人家。于是，主人家把水在路旁撒了十几分钟，才把水送到主人家的院子里。这天，主人又让他们一起去挑运水回来。他们一趟挑运回来，都完好无损，就这样过了十几天。最后，他们又去挑运回来的那十几桶水。这下，他们俩可以再去路旁采一桶水了。从此，他们两人再也没有过半桶的水了，再也没有碰到过别的挑运来采水的了。 \\ 
 Once upon a time, there was a water-bearer who wanted to deliver half a bucket of water sprinkled by the roadside to the master's house. The master asked him to send the water the broken bucket noticed sprinkled by the roadside to the master's house. So, the master sprinkled water on the roadside for more than ten minutes, until the water was delivered to the master's yard. On this day, the master asked them to carry water together again. They came back from a single trip, and all trips were all intact, and it was like this for more than ten days. Finally, they went to pick up the dozen or so buckets of water that were brought back. Now, the two of them can go to the roadside to get a bucket of water. Since then, the two of them have never had more than half a bucket of water, and have never encountered other people who came to collect water.\\

\midrule
\textbf{\textbf{- w/ Semantics}} \quad \small 有一个挑水夫因为完好无损，只能剩下半桶水。挑水人问他为什么不把水送到主人家，主人说，这趟挑运过去，只用了三秒钟，就把水运过去了。没多久，挑水的人就饿死了！主人对他道歉，并把那桶水送到了路旁撒了一次又一次网。最后，那人得救了，也终于和主人家的家长们道了歉。
\\
There was a water-bearer who had only half a bucket of water left because it was intact. The water-bearer asked him why he didn't deliver the water to the master's house. The master said that this trip took only three seconds to deliver the water. Not long after, the water-bearer starved to death! The master apologised to him and sent the bucket of water to the roadside for casting the net again and again. In the end, the man was saved, and he finally apologised to the parents of the master.\\
\midrule
\textbf{Ours} \quad \small 从前有个挑水夫，每天都要把水送到主人家。挑水的人觉得，这趟挑运真是完好无损。于是，他就去路旁捡了一个破桶，希望能够用它把水运过去。结果破桶里的水被撒了个精光。主人家知道后，非常生气。然后就下令把那个挑水人淹死在路旁。 \\ 
Once upon a time, there was a water-bearer who deliver water to the master's house every day. Water-bearer felt that the trip was really intact. So he went to the roadside to pick up a broken bucket, hoping to use it to deliver the water. As a result, the water in the broken bucket was sprinkled all over. When the master found out, he was very angry. Then, the master ordered the water-bearer to be drowned by the roadside.\\
\hline
\bottomrule[1pt]
\end{tabular}
\caption{The given outline and corresponding generated stories for the case study.}
\label{tab:case_study}
\end{center}
\end{table*}

\end{CJK}
\end{document}